\documentclass{article}

\usepackage[preprint]{corl_2022} 

\usepackage{amsmath}
\usepackage{amssymb}
\usepackage{bbm}
\usepackage{booktabs}
\usepackage{combelow}
\usepackage{epsfig}
\usepackage{graphicx}
\usepackage{multirow}
\usepackage{pifont}
\usepackage{subfig}
\usepackage{times}
\usepackage[table]{xcolor}

\newcommand{\cmark}{\ding{51}}
\newcommand{\xmark}{\ding{55}}

\title{When the Sun Goes Down: Repairing Photometric Losses for All-Day Depth Estimation}

\author{
	Madhu Vankadari \\
	University of Oxford \\
	\texttt{madhu.vankadari@cs.ox.ac.uk} \\
	\And
	Stuart Golodetz \\
	University of Oxford \\
	\texttt{stuart.golodetz@cs.ox.ac.uk} \\
	\And
	Sourav Garg \\
	Queensland University of Technology \\
	\texttt{s.garg@qut.edu.au} \\
	\And
	Sangyun Shin \\
	University of Oxford \\
	\texttt{sangyun.shin@cs.ox.ac.uk} \\
	\And
	Andrew Markham \\
	University of Oxford \\
	\texttt{andrew.markham@cs.ox.ac.uk} \\
	\And
	Niki Trigoni \\
	University of Oxford \\
	\texttt{niki.trigoni@cs.ox.ac.uk}
}

\begin{document}
\maketitle

\vspace{-\baselineskip}

\begin{abstract}
Self-supervised deep learning methods for joint depth and ego-motion estimation can yield accurate trajectories without needing ground-truth training data. However, as they typically use photometric losses, their performance can degrade significantly when the assumptions these losses make (e.g.\ temporal illumination consistency, a static scene, and the absence of noise and occlusions) are violated. This limits their use for e.g.\ nighttime sequences, which tend to contain many point light sources (including on dynamic objects) and low signal-to-noise ratio (SNR) in darker image regions. In this paper, we show how to use a combination of three techniques to allow the existing photometric losses to work for both day and nighttime images. First, we introduce a per-pixel neural intensity transformation to compensate for the light changes that occur between successive frames. Second, we predict a per-pixel residual flow map that we use to correct the reprojection correspondences induced by the estimated ego-motion and depth from the networks. And third, we denoise the training images to improve the robustness and accuracy of our approach. These changes allow us to train a single model for both day and nighttime images without needing separate encoders or extra feature networks like existing methods. We perform extensive experiments and ablation studies on the challenging Oxford RobotCar dataset to demonstrate the efficacy of our approach for both day and nighttime sequences.
\end{abstract}

\section{Introduction}
\label{sec:intro}

An ability to capture 3D scene structure is crucial for many applications, including autonomous driving~\cite{yurtsever2020survey}, robotic manipulation~\cite{sanchez2018robotic}, and augmented reality~\cite{livingston2009indoor}. Many methods use LiDAR or fixed-baseline stereo to acquire the depth needed to reconstruct a scene, but researchers have also long been interested in estimating depth from monocular images, driven by the ubiquity, low cost, low power consumption and ease of deployment of monocular cameras. By contrast, LiDAR can be power-hungry, and stereo rigs must be calibrated and time-synchronised to achieve good performance.

Multi-view monocular depth estimation approaches have long used variable-baseline stereo over multiple images to recover depth \cite{newcombe2011dtam, wang2018mvdepthnet}. Meanwhile, progress in deep learning has opened up the additional possibility of estimating depth from a single monocular image. Deep learning methods for depth estimation can be broadly divided into two types, namely supervised methods \cite{eigen2014depth,liu2015deep}, and self/unsupervised methods \cite{garg2016unsupervised,zhan2018unsupervised,almalioglu2018ganvo}. Typically, supervised approaches have achieved very good results for the dataset(s) on which they are trained, but their need for ground-truth information during training has often hindered their deployment in new domains.
   
\begin{figure}[!t]
\centering
\includegraphics[scale=0.185]{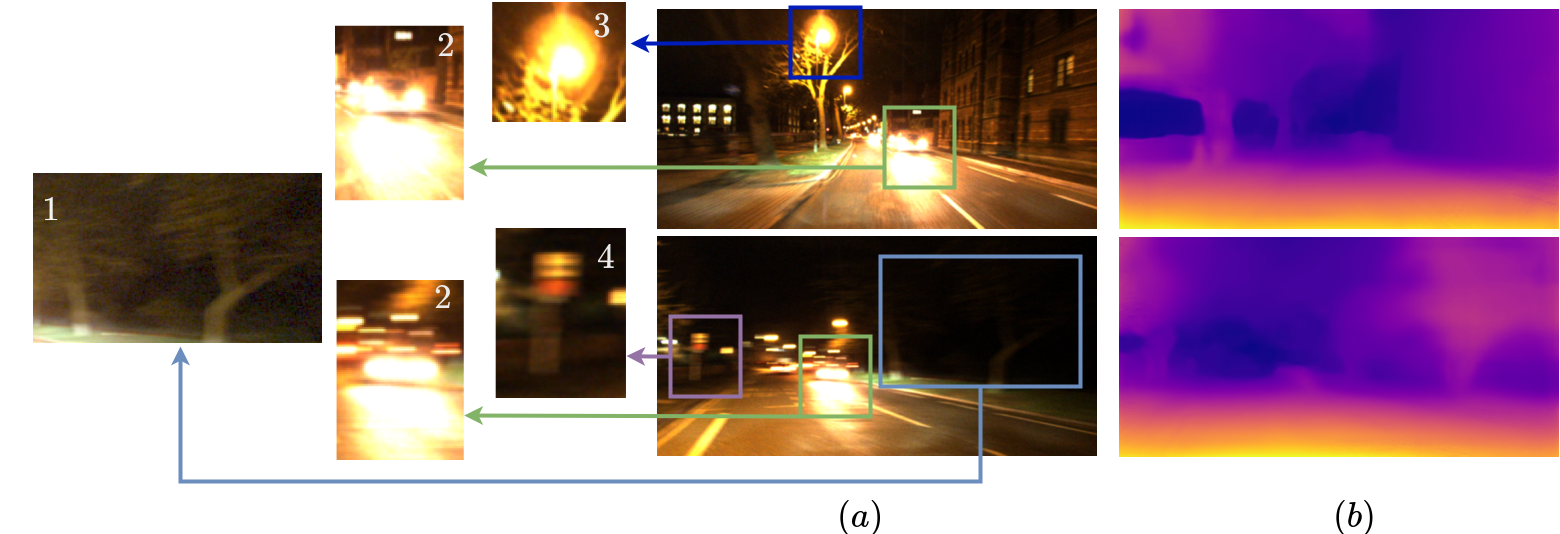}
\caption{(a) The challenges posed by night-time images: (1) low visibility and noise (patch enhanced for better readability); (2) moving light sources with saturating image regions; (3) point light sources; (4) extreme motion blur. (b) Despite these adverse conditions, which violate the assumptions made by the photometric loss, our method can successfully estimate accurate depth maps.}
\label{fig:challenges}
\vspace{-\baselineskip}
\end{figure}

By contrast, self/unsupervised methods have typically adopted the use of a geometry-based loss function, inspired by the strong physical principles of traditional methods~\cite{kerl2013dense,kerl2013robust}. This loss function is commonly referred to as the photometric or appearance loss, and is based on the assumptions that (i) the scene is static (i.e.\ contains no moving objects), (ii) the illumination in the scene is diffusive (i.e.\ there are no specular reflections) and temporally consistent (i.e.\ the pixels to which any scene point projects in any two consecutive frames have the same intensity), and (iii) the images are free of noise and occlusions~\cite{kerl2013dense, kerl2013robust, comport2007accurate, sharma2020nighttime}. In practice, many of these assumptions are at least partly false, which can lead to errors in the estimated depth: scenes are quite likely to contain dynamic objects (e.g. cars, cyclists and pedestrians, in an outdoor driving scenario), surface materials are rarely fully diffusive, and occlusions are common. During the day, it is somewhat reasonable to assume that the illumination is moderately temporally consistent for image sequences captured outdoors, as the sun is by far the dominant light source in that case, and the light it casts changes only slowly over time; however, at night, the numerous point light sources that are typically turned on after dark (e.g.\ car headlights, lamp posts, etc.) can cause the illumination to change drastically from one frame to the next. At night, also, the motion blur associated with the movement of dynamic objects in the scene (including the ego-vehicle) becomes worse, owing to the longer exposure times typically used when capturing night-time images \cite{portz2012optical,rav2000restoration}, and the signal-to-noise ratio of the (darker) images becomes much lower than it would be during daytime. Such issues, as illustrated in Figure~\ref{fig:challenges}, inhibit the straightforward use of deep networks based on photometric loss for night-time sequences.

In this paper, we address this problem by directly targeting violations of the temporal illumination consistency, static scene and noise-free assumptions on which the photometric loss relies. As shown by our day and night results in Table~\ref{tab:depth_results}, these three together account for much of the discrepancy in performance between daytime and night-time. A lack of temporal illumination consistency caused by point light sources in the scene can cause pixels to be incorrectly matched between consecutive frames. To rectify this, we propose a novel per-pixel neural intensity transformation that learns to compensate for these light sources (see \S\ref{sec:NIT}). Whilst conceptually straightforward, this approach is surprisingly effective, as our results in \S\ref{sec:experiments} demonstrate. Interestingly, they also show that it is able to operate well over wide (motion parallax) baselines, allowing us to leverage the better depth estimation performance that wider baselines offer. To correct for dynamic objects in the scene, as well as motion blur, we predict a per-pixel residual flow map (see \S\ref{subsec:residuals}) that we use to correct the reprojection correspondences induced by the estimated ego-motion and depth from the networks. This improves depth estimation performance at any time of day (see \S\ref{sec:experiments}), but has additional theoretical benefits for night-time sequences because of the greater motion blur from which they typically suffer. Lastly, we robustify our approach against noise by incorporating Neighbour2Neighbour~\cite{huang2021neighbor2neighbor}, a state-of-the-art denoising module, in our photometric loss formulation (see \S\ref{sec:denoising}).

\section{Related Work}
\label{sec:related_work}

Estimating depth from images has a long history in computer vision. Several methods use either stereo images~\cite{scharstein2002taxonomy,scharstein2007learning,zou2010method}, or two or more images taken from different viewing angles~\cite{schonberger2016structure,dai2013projective,yu20143d}. We try to solve this problem using a single  monocular image, without any constraints on the scene of interest. Various methods have addressed this problem using supervised learning~\cite{eigen2014depth,liu2015deep,ladicky2014pulling,dos2019sparse,fu2018deep}. However, it is infeasible to have ground-truth depth maps for training on every scene, which limits the application of these methods and helps motivate unsupervised solutions to this problem.

\textbf{Unsupervised Methods:} Garg et al.~\cite{garg2016unsupervised} proposed a geometry-based loss function to train a network in a completely unsupervised fashion using a pair of stereo images. Monodepth~\cite{godard2017unsupervised} improved this by using differentiable image warping~\cite{jaderberg2015spatial} and structural similarity-based~\cite{wang2004image} image comparison loss. SfMLearner~\cite{zhou2017unsupervised} used only monocular images to jointly learn depth and ego-motion. It was further improved by combining stereo and monocular losses in~\cite{babu2018undemon,li2018undeepvo}. Later, GeoNet~\cite{yin2018geonet} and EPC~\cite{luo2018every} learnt per-pixel optical flow maps along with depth and ego-motion to mitigate the effect of moving objects. Some methods use GAN-based learning to train their systems~\cite{almalioglu2018ganvo,aleotti2018generative,vankadari2019unsupervised}. Recently, Monodepth2~\cite{godard2018digging} extended Monodepth to the temporal domain, proposing a few architectural changes and robust loss functions to achieve state-of-the-art results. HR-Depth~\cite{lyu2021hr} used an effective skip connection and a convolution block to integrate spatial and semantic information. SD-SSMDE~\cite{petrovai2022exploiting} introduced a two-stage training strategy to improve scale and inter-frame scale consistency in depth by utilising depth estimation from the first stage as a pseudo-label. Based on channel-wise attention, CADepth-Net~\cite{yan2021channel} proposed structure perception and detail emphasis modules for capturing the context of scenes with the detail for the depth estimation. RM-Depth~\cite{hui2022rm} proposed recurrent modulation units for an effective fusion of deep features with fewer parameters, and a warping-based motion field for moving objects to improve the scene rigidity, leading to enhanced depth estimation. More broadly, recent years have also seen a wide range of other advances in depth estimation, e.g.\ changes to the network architecture~\cite{guizilini20203d}, the addition of extra loss functions~\cite{shu2020feature}, and better handling of dynamic objects~\cite{li2020unsupervised}. However, all of these methods have been tested on standard daytime datasets, whereas our method is designed to work at night as well.

\textbf{Nighttime Methods:} All the methods above are trained using photometric loss as the main supervision signal, and with an assumption of temporal illumination consistency, which is not valid at night. A few methods, such as DeFeat-Net~\cite{spencer2020defeat}, ADFA~\cite{vankadari2020unsupervised} and~\cite{sharma2020nighttime}, have explored how to estimate depth information from nighttime RGB images. DeFeat-Net~\cite{spencer2020defeat} learns $n$-dimensional deep feature representations (assumed to be illumination-invariant) using a pixel-wise contrastive loss. The feature maps are simultaneously used along with the images for photometric loss calculation during training. ADFA~\cite{vankadari2019unsupervised} mimics a daytime depth estimation model by learning a new encoder that can generate `day-like' features from nighttime images using a domain adaptation approach. Instead of feature translation as in~\cite{vankadari2020unsupervised}, the authors in~\cite{sharma2020nighttime} propose a joint network for image translation and stereo image-based depth estimation. Recently, photometric losses are again used with an image enhancement module and a GAN-based depth regulariser in \cite{wang2021regularizing}. Liu et al.~\cite{liu2021self} divided the day and nighttime images into view-invariant and variant feature maps using separate encoders, and used the view-invariant information for depth estimation. All these methods either need two separate encoders for day and nighttime images~\cite{vankadari2020unsupervised, liu2021self,wang2021regularizing}, or need to learn an illumination-invariant feature space~\cite{spencer2020defeat}. By contrast, our proposed method learns in a completely self-supervised fashion, without needing stereo images, ground-truth depth information or any additional feature learning.

\section{Method}
\label{sec:method}

\begin{figure}
\centering
\includegraphics[scale=0.28]{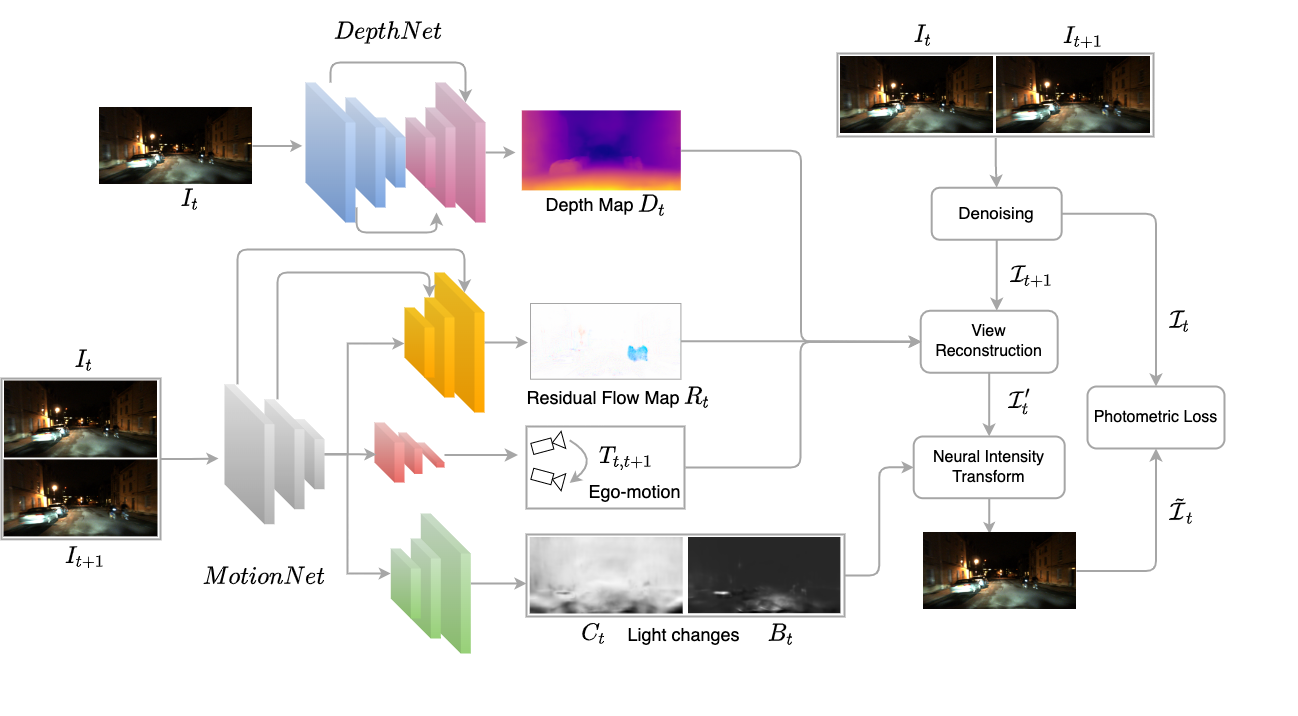}
\vspace{-\baselineskip}
\caption{The architecture of our proposed method (see \S\ref{sec:method} for details).}
\label{fig:arch_diag}
\vspace{-\baselineskip}
\end{figure}

\subsection{Baseline Method}
\label{sec:baseline}

We first recap the core tenets of existing photometric loss methods, which typically use two networks, a depth network (or \emph{DepthNet}) and a motion network (or \emph{MotionNet}). The \emph{DepthNet} takes an individual colour image as input, and is used to predict a depth image $D_t$ for each colour image $I_t$ in the input sequence. The \emph{MotionNet} takes a consecutive pair of images $I_t$ and $I_{t+1}$ as input, and is used to output the ego-motion $T_{t,t+1}$ of the camera between them. The estimated depth and ego-motion can be used to reproject a pixel $\mathbf{u} = [u,v]^\top$ in frame $I_t$ into $I_{t+1}$, the subsequent frame in the sequence, via $\dot{V_t}(\mathbf{u}) = KT_{t,t+1}D_t(\mathbf{u})K^{-1}\dot{\mathbf{u}}$, in which $\dot{\mathbf{u}}$ denotes the homogeneous form of $\mathbf{u}$, $K \in \mathbb{R}^{3 \times 3}$ encodes the camera intrinsics, and $\dot{V_t}(\mathbf{u}) \in \mathbb{R}^3$ denotes the homogeneous form of $V_t(\mathbf{u}) \in \mathbb{R}^2$, a 2D point in the image plane of $I_{t+1}$ (which may or may not lie within the bounds of the actual image). This can be used to reconstruct an image $I'_t$ by sampling from $I_{t+1}$ around the reprojected points, using bilinear interpolation \cite{jaderberg2015spatial} to achieve a smoother result. Formally,
\begin{equation}
\textstyle I'_t(\mathbf{u}) = \begin{cases}
\mbox{interpolate}(I_{t+1}, V_t(\mathbf{u})) & \mbox{if } \mathbf{u} \in M_t \\
\mathbf{0} & \mbox{otherwise},
\end{cases}
\label{eqn:recon}
\end{equation}
in which $M_t = \{\mathbf{u} : \rho(V_t(\mathbf{u})) \in \Omega(I_{t+1})\}$ is the set of pixels whose reprojections into $I_{t+1}$, when rounded to the nearest pixel using $\rho$, fall within the image bounds $\Omega(I_{t+1})$. The reconstructed image $I'_t$ can then be compared to the original image $I_t$ to calculate the loss values needed for training. The loss we target, namely \emph{photometric loss}, has been used by many recent deep learning-based depth estimation techniques \cite{garg2016unsupervised,babu2018undemon,zhou2017unsupervised,vankadari2019unsupervised}. It is normally calculated as a convex combination of pixel-wise difference and single-scale structural dissimilarity (SSIM) \cite{wang2004image}, via
\begin{equation}
\textstyle L_p^{(t)} = \frac{1}{|M_t|} \sum_{\mathbf{u} \in M_t} \left( \alpha \frac{1 - \mathit{SSIM}(I_t(\mathbf{u}), I'_t(\mathbf{u}))}{2} + (1 - \alpha) \left| I_t(\mathbf{u}) - I'_t(\mathbf{u}) \right| \right).
\end{equation}
Most existing unsupervised methods (e.g.\ \cite{zhou2017unsupervised,babu2018undemon,godard2018digging,guizilini20203d}) use this as the backbone of their formulation. To ensure a fair comparison with current night-time state-of-the-art methods \cite{spencer2020defeat, liu2021self,wang2021regularizing}, we base our modifications in this paper on Monodepth2 \cite{godard2018digging}, a commonly used baseline.

\subsection{Lighting Change Compensation}
\label{sec:NIT}

The numerous point light sources that are typically turned on after dark (e.g.\ car headlights, lamp posts, etc.) can cause the illumination of a scene to change significantly from frame $I_t$ to frame $I_{t+1}$. A problematic special case occurs when a light source moves with the camera (e.g.\ car headlights), which can lead to large holes in the estimated depth directly in front of the ego-vehicle \cite{spencer2020defeat, liu2021self}. In our approach, we compensate for the illumination changes by estimating a per-pixel transformation that, when applied to $I_{t+1}$, can mitigate the changes in lighting that have occurred since $I_t$. We draw some inspiration from \cite{jin2001real,baker2004lucas,yang2020d3vo}, which use a single whole-image transformation based on two scalar values to compensate for the difference in exposure time between a pair of images, based on the observation that such a difference creates approximately uniform intensity changes over the entire image. However, in our case, the intensity changes are far from uniform over the image, owing to both the motions of the ego-vehicle and other objects in the scene, and the distances between the ego-vehicle and static point light sources. For this reason, we propose a per-pixel formulation here.

Our approach starts by passing the features produced by the last convolutional layer of the \emph{MotionNet} through a \emph{lighting change decoder} to estimate two per-pixel change images, $C_t$ and $B_t$ (see Figure~\ref{fig:arch_diag}). These (respectively) aim to capture the per-pixel changes in contrast (scale) and brightness (shift) that have occurred between the two input frames. As shown in Figure~\ref{fig:ablation}, the brightness image $B_t$ broadly captures the extra light added to the image by e.g.\ vehicle headlights, and the contrast image $C_t$ broadly captures the changes in ambient light due to the motion of the ego-vehicle towards or away from point light sources such as street lamps. We use these images to transform the reconstructed image $I'_t$ via $\tilde{I}_t = C_t \odot I'_t + B_t$, in which $\odot$ denotes the Hadamard product.

\subsection{Motion Compensation}
\label{subsec:residuals}

As seen in \S\ref{sec:baseline}, the standard photometric loss makes use of correspondences between consecutive frames that have been established via reprojection, based on the ego-motion and depth estimated by the networks. Assuming that (i) the ego-motion and depth have been estimated well, (ii) the scene is static, and (iii) there is minimal motion blur, the correspondences established in this way will broadly match those that would have been established had we used the ground truth optic flow $\Phi_t(\cdot)$ from frame $t$ to frame $t+1$. However, if objects move with respect to the background scene, or anything visible in the image moves with respect to the ego-camera (which can cause motion blur), then the reprojection correspondences may be incorrect. To correct for these errors, we predict a residual flow map $R_t$, such that for each pixel $\mathbf{u} \in \Omega(I_t)$, $R_t(\mathbf{u}) \in \mathbb{R}^2$ is an estimate of $(\mathbf{u} + \Phi_t(\mathbf{u})) - V_t(\mathbf{u})$, the 2D offset from the reprojection correspondence of $\mathbf{u}$, namely $V_t(\mathbf{u})$, to its ground truth correspondence in frame $t+1$, namely $\mathbf{u} + \Phi_t(\mathbf{u})$. We can then add $R_t(\mathbf{u})$ to $V_t(\mathbf{u})$ for each pixel $\mathbf{u}$ to obtain a potentially more accurate correspondence for use in reconstructing $I'_t$ via Equation~\ref{eqn:recon}.

Some methods \cite{luo2018every, yin2018geonet, li2020unsupervised} already exist that predict residual flow for daytime images. By contrast, we avoid using a separate encoder-decoder network or computationally-intensive image warping-based bilinear interpolation for supervision. Instead, we estimate residual flow using an efficient sparsity-based formulation.
This involves introducing a \emph{residual flow decoder} that takes the features of the final convolutional layer of the \emph{MotionNet} as input and the features of previous layers in the \emph{MotionNet} via skip connections, and outputs residual flow maps $\{R_{t,s} : s \in \{0,1,2,3\}\}$ at four different scales (each $R_{t,s}$ has a width and height that is $1/2^s$ that of $I_t$, and $R_{t,0} \equiv R_t$).

There is no direct supervision available to learn the residual flow maps. For this reason, we choose instead to encourage sparsity in the residual flow estimates, so that the estimated depth and ego-motion can explain the majority of the scene, and the left-over can be explained by the residual flow maps. To achieve this, we adopt the sparsity loss from~\cite{li2020unsupervised}, i.e.
\begin{equation}
\textstyle L_r^{(t)} = \sum_{s=0}^3 \langle|R_{t,s}|\rangle / 2^s \sum_{\mathbf{u} \in \Omega(I_{t,s})} \sqrt{1+|R_{t,s}(\mathbf{u})| / \langle|R_{t,s}|\rangle},
\end{equation}
in which $I_{t,s}$ is a downsampled version of $I_t$ at scale $s$, and $\langle|R_{t,s}|\rangle$ is the spatial average of the absolute residual flow map $|R_{t,s}|$. By contrast with \cite{li2020unsupervised}, here we introduce a normalising factor of $1/2^s$ at each scale, since the original loss was for scene flow, where the flow magnitude is independent of the resolution of the flow maps, which is not the case for the 2D residual flow we consider.

\subsection{Image Denoising}
\label{sec:denoising}

Image noise is yet another key factor that affects the performance of the photometric loss. In practice, it is independent of the respective image, and is mainly caused by a low SNR in the darker regions of the image. Handling this noise is of crucial importance, as photometric loss is the only training signal, and supervises all of the modules we have mentioned thus far. To remove the noise from the images, we chose to use Neighbour2Neighbour~\cite{huang2021neighbor2neighbor}, a state-of-the-art unsupervised denoising model trained on ImageNet with zero-mean Gaussian noise. The standard deviation values were varied from $5$ to $50$ during training. This model can either be used to denoise all images input to the network at both training time and test time, or it can be used solely at training time to denoise the images for the purpose of calculating the loss. In practice, we chose the latter approach, as denoising at test time has two major disadvantages: (i) it can significantly add to the computational burden at runtime, slowing down the depth estimation; and (ii) any errors in the denoising process can lead to downstream errors in the depth maps, even though the depth estimation model itself might have been trained well. By contrast, restricting denoising to training time has the advantage of allowing us to make the depth and motion networks robust to noise by training them on the original images.

\subsection{Full Pipeline}

We can now formulate our full pipeline as follows:
\begin{equation}
\footnotesize
\begin{aligned}
\textstyle D_t &= \textstyle \mathcal{D}(I_t), \; f_n = \mathcal{ME}_{1:n}([I_t, I_{t+1}]) \\
\textstyle T_{t,t+1} &= \textstyle \mathcal{MD}(f_N), \; R_t = \mathcal{RFD}(\{f_n : 1 \le n \le N\}), \; (C_t, B_t) = \mathcal{LCD}(f_N) \\
\textstyle \mathcal{I}_t &= \textstyle \mathcal{DN}(I_t), \; \mathcal{I}_{t+1} = \mathcal{DN}(I_{t+1}) \\
\textstyle \mathcal{I}'_t &= \textstyle \mbox{reconstruct}(\mathcal{I}_{t+1}, V_t + R_t) \\
\textstyle \mathcal{\tilde{I}}_t &= \textstyle C_t \odot \mathcal{I}'_t + B_t \\
\textstyle L_p^{(t)} &= \textstyle \frac{1}{|M_t|} \sum_{\mathbf{u} \in M_t} \left( \alpha \frac{1 - \mathit{SSIM}(\mathcal{I}_t(\mathbf{u}), \mathcal{\tilde{I}}_t(\mathbf{u}))}{2} + (1 - \alpha) \left| \mathcal{I}_t(\mathbf{u}) - \mathcal{\tilde{I}}_t(\mathbf{u}) \right| \right)
\end{aligned}
\end{equation}
The inputs to our system are a consecutive pair of images $I_t$ and $I_{t+1}$, whilst $\mathcal{D}$ denotes the \emph{DepthNet}, $\mathcal{ME}_{1:n}$ denotes the first $n$ layers of the $N$-layer \emph{MotionNet} encoder, $\mathcal{MD}$ denotes the \emph{MotionNet} decoder, $\mathcal{RFD}$ denotes the residual flow decoder, $\mathcal{LCD}$ denotes the lighting change decoder, and $\mathcal{DN}$ denotes the denoiser \cite{huang2021neighbor2neighbor}. The \texttt{reconstruct} function reconstructs $\mathcal{I}'_t$ as per Equation~\ref{eqn:recon}.

\subsection{Making the Pipeline Bidirectional}

Monodepth2~\cite{godard2018digging} calculates its photometric loss not only in the forwards direction, from $I_t$ to $I_{t+1}$, but also in the backwards direction, from $I_t$ to $I_{t-1}$, before combining the losses. This allows us to use the idea of minimum reprojection error to account for occluded pixels, and so we do the same. We also adopt the auto-masking losses $L_a^{(t)}$ from Monodepth2~\cite{godard2018digging}, as even though our method can cope with moving objects, it is very difficult to use parallax to disentangle the motion of objects that are moving in the same direction and at the same speed as the ego-vehicle. We further include the commonly used edge-aware gradient smoothing loss $L_g^{(t)}$ \cite{godard2018digging} to maintain spatial smoothness over the estimated depth maps. Our final loss $L^{(t)}$ then becomes the weighted sum
\begin{equation}
    \textstyle L^{(t)} = \min\left(L_{p-}^{(t)}, L_{p+}^{(t)},L_{a-}^{(t)},L_{a+}^{(t)}\right) + \lambda_r \left(L_{r-}^{(t)} + L_{r+}^{(t)}\right) + \lambda_g L_g^{(t)},
\end{equation}
in which $+$/$-$ denote the forward/backward versions of the losses, and $\lambda_r, \lambda_g \in \mathbb{R}$ are the weights.

\section{Experiments}
\label{sec:experiments}

In \S\ref{subsec:depthevaluation}, we compare our depth estimation performance to a number of state-of-the-art approaches in a variety of different daytime and/or night-time contexts. In \S\ref{suppSec:parallax}, we present a study on the effect of parallax to help explain the importance of our neural intensity transformation module. Finally, in \S\ref{subsec:ablationstudies}, we perform an ablation study to analyse the contributions made by the three individual components of our approach. Further experiments can be found in the supplementary material.

\subsection{Depth Evaluation}
\label{subsec:depthevaluation}

\begin{figure*}[!t]
\centering
\includegraphics[scale=0.24]{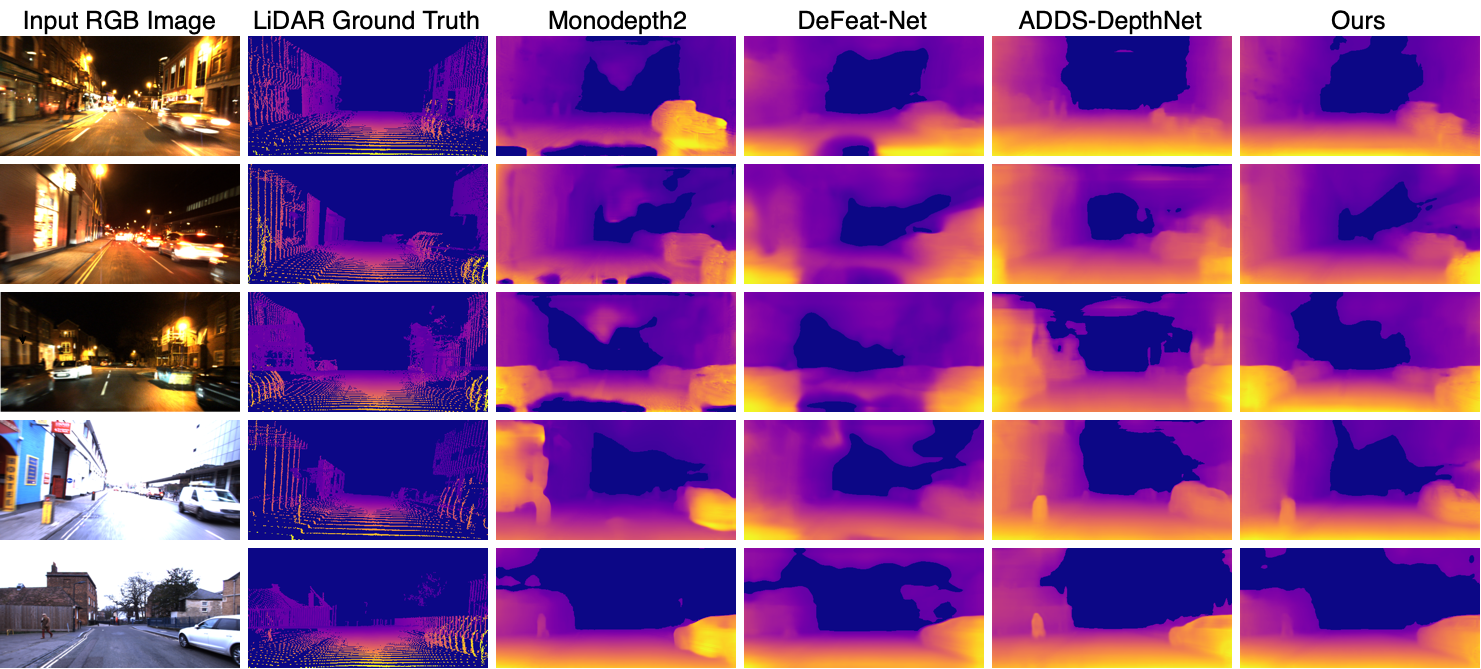}
\caption{A qualitative comparison of our proposed method with the state of the art.}
\label{fig:depth_results}
\vspace{-\baselineskip}
\end{figure*}

We compare with $4$ state-of-the-art unsupervised monocular methods: Monodepth2~\cite{godard2018digging},  DeFeat-Net~\cite{spencer2020defeat}, ADDS-Depth-Night~\cite{liu2021self} and RNW~\cite{wang2021regularizing} (see Figure~\ref{fig:depth_results} and Table~\ref{tab:depth_results}). We tested our model with 3 different data variations: day only ($d$), night only ($n$), and a mix of day and night ($d\&n$). Monodepth2~\cite{godard2018digging} can be trained with all 3 configurations, although it has already been outperformed by DeFeat-Net~\cite{spencer2020defeat} in the $d\&n$ setting. For the $d$ and $n$ settings, we outperform it by a significant margin in both error and accuracy (see Table~\ref{tab:depth_results}). DeFeat-Net~\cite{spencer2020defeat} and ADDS-Depth-Night~\cite{liu2021self} were originally trained with a $d\&n$ configuration. We evaluated the pre-trained models they released on our test split. Our method outperforms both methods by a significant margin on the nighttime sequences (see Table~\ref{tab:depth_results}). Please note that we do not use any additional feature representation-based losses as used in DeFeat-Net~\cite{spencer2020defeat}, or paired day and night images as used in ADDS-Depth-Net~\cite{liu2021self}. RNW~\cite{wang2021regularizing}, another recent method, is also built on Monodepth2, but targets nighttime data only. As per Figure~\ref{fig:depth_results}, our depth estimation results are sharp and better able to preserve edges than the competing methods. We also found that using a longer baseline improves depth estimation performance. However, na\"{i}vely using a wider baseline without also using our neural intensity transform can lead to a severe decrease in accuracy, particularly for nighttime images.

\begin{table*}[!t]
\begin{center}
\resizebox{\textwidth}{!}{
\begin{tabular}{cccccccccc}
\toprule
\textit{Test} & \textit{Method} & \textit{Train} & \textit{Abs.\ Rel.} & \textit{Sq. Rel.} & \textit{RMSE} & \textit{Log RMSE} & $\delta < 1.25$ & $\delta < 1.25^2$ & $\delta< 1.25^3$ \\
\midrule

\multirow{6}{*}{\rotatebox{90}{\textit{Day}}} & Monodepth2 \cite{godard2018digging} & d & 0.219 & 4.525 & 7.641 & 0.285 &  0.679 & 0.862 & 0.930 \\
& Ours & d & \textbf{0.191} & \textbf{1.710} & \textbf{6.158} & \textbf{0.253} & \textbf{0.713} & \textbf{0.904} &  \textbf{0.962} \\
\cmidrule{2-10}
& DeFeat-Net~\cite{spencer2020defeat} & d \& n & 0.247 & 2.980 & 7.884 & 0.305 & 0.650 & 0.866 & 0.943 \\
& RNW~\cite{wang2021regularizing} & \underline{d \& n} & 0.297 & 2.608 & 7.996 & 0.359 & 0.431 & 0.773 & 0.930 \\
& ADDS-Depth-Night~\cite{liu2021self} & \underline{d \& n} & 0.239 & 2.089 & 6.743 & 0.295 & 0.614 & 0.870 & 0.950 \\
& Ours & d \& n & \textbf{0.176} & \textbf{1.603} & \textbf{6.036} & \textbf{0.245} & \textbf{0.750} & \textbf{0.912} & \textbf{0.963} \\

\midrule

\multirow{7}{*}{\rotatebox{90}{\textit{Night}}} & Monodepth2~\cite{godard2018digging} & n & 0.453 & 21.310 & 11.420 & 0.444 & 0.700 & 0.873 & 0.930 \\
& RNW MCIE + SBM~\cite{wang2021regularizing} & n & 0.350 & 7.934 & 8.994 & 0.407 & 0.674 & 0.861 & 0.922 \\
& Ours & n & \textbf{0.186} & \textbf{1.656} & \textbf{6.288} & \textbf{0.248} & \textbf{0.728} & \textbf{0.919} & \textbf{0.969} \\
\cmidrule{2-10}
& DeFeat-Net~\cite{spencer2020defeat} & d \& n & 0.334 & 4.589 & 8.606 & 0.358 & 0.586 & 0.827 & 0.911 \\
& ADDS-Depth-Night~\cite{liu2021self} & \underline{d \& n} & 0.287 & 2.569 & 7.985 & 0.339 & 0.490 & 0.816 & 0.946 \\
& RNW~\cite{wang2021regularizing} & \underline{d \& n} & 0.185 & 1.710 & 6.549 & 0.262 & 0.733 & 0.910 & 0.960 \\
& Ours & d \& n & \textbf{0.174} & \textbf{1.637} & \textbf{6.302} & \textbf{0.245} & \textbf{0.754} & \textbf{0.915} & \textbf{0.964} \\
\bottomrule
\end{tabular}
}
\end{center}
\vspace{-.5\baselineskip}
\caption{A quantitative comparison of our method. The results of Monodepth2 \cite{godard2018digging} are reported after retraining it. Those of DeFeat-Net~\cite{spencer2020defeat} and ADDS-Depth-Night~\cite{liu2021self} are reported using the checkpoints from their public repositories. The evaluation uses a maximum depth of $50$m. Underlined methods use daytime images as main supervision or for regularisation losses.}
\label{tab:depth_results}
\vspace{-\baselineskip}
\end{table*}

\subsection{Effect of Parallax}
\label{suppSec:parallax}

To better understand how depth estimation performance is affected by increasing the average parallax (metric separation) between the images we use to calculate the photometric loss, we constructed a new nighttime training split by increasing the intra-triplet stride (see supplementary material) to $2$, which increased the average parallax between the images from $0.353$m to $0.706$m. Without our neural intensity transformation, the depth estimation performance significantly decreased compared to the original nighttime training split (see the difference between the RMSEs of the baseline in the top and bottom parts of Table~\ref{tab:ablation}). A key cause of this in night images is likely the headlights of the ego-vehicle, which can cause the pixel intensities to change drastically between frames. However, with our neural intensity transformation, the depth estimation performance was found to instead increase, which we hypothesise to be because by compensating for the lighting changes, we make it possible to exploit the stronger supervision that can be offered by a wider baseline.

\subsection{Ablation Study}
\label{subsec:ablationstudies}

\begin{figure}[!t]
\centering
\includegraphics[scale=0.27]{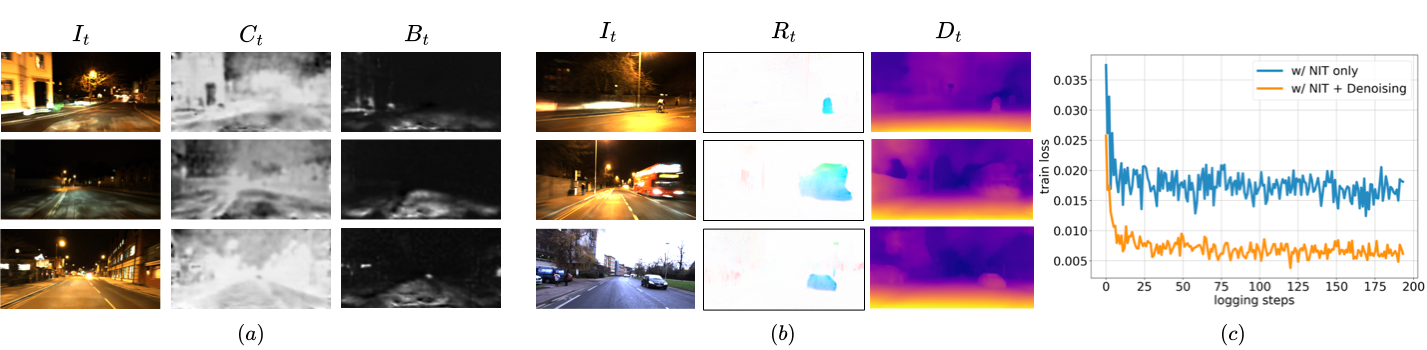}
\caption{Visualisations of (a) estimated light changes; (b) residual flow and estimated depth; (c) the effects of denoising on the training loss over time.}
\label{fig:ablation}
\vspace{-\baselineskip}
\end{figure}

\textbf{Lighting Change Compensation.} In Figure~\ref{fig:ablation}(a), we show several reference images and their lighting change maps. The intensity changes are non-uniform, so we cannot use the existing correction approaches from \cite{jin2001real,baker2004lucas,yang2020d3vo}. We also observe that our method is able to clearly disentangle both the changes in ambient light resulting from movement towards/away from point light sources (captured by $C_t$) and the additional light added to the road pixels in the images by the ego-vehicle headlights (captured by $B_t$). Our neural intensity transform significantly reduces the RMSE error compared to the baseline (see Table~\ref{tab:ablation}), and is also able to fill in holes in front of the ego-vehicle (see Figure~\ref{fig:depth_results}).

\textbf{Motion Compensation.} In Figure~\ref{fig:ablation}(b), we show several reference images and their residual flow and depth maps. In the second column, one can clearly see that our method is able to distinguish pixels on moving objects such as cars and pedestrians from static pixels. This effect can be observed for both daytime and nighttime images, showcasing the generality of our approach through a single unified training pipeline. In Table~\ref{tab:ablation}, it can be seen that correcting the reprojection correspondences using the residual flow map we predict leads to a significant improvement in accuracy.

\textbf{Image Denoising.} Denoising the images while calculating the training loss should ideally reduce the ambiguity in establishing pixel correspondences between the images, giving a robust supervision signal for training our system and thereby achieving lower error and higher accuracy. This effect can be clearly seen in the training error plot shown in Figure~\ref{fig:ablation}(c), where we compare our baseline+NIT model with and without denoising. The denoising results in much more accurate depth maps, improving both the RMSE and accuracy metrics as shown in Table~\ref{tab:ablation}.

\begin{table}[!t]
\centering
\scriptsize
\begin{tabular}{ccccccccc}
\toprule
\textit{Stride} & \textit{Method} & \textit{Abs.\ Rel.} & \textit{Sq.\ Rel.} & \textit{RMSE} & \textit{Log RMSE} & $\delta < 1.25$ & $\delta < 1.25^2$ & $\delta< 1.25^3$ \\
\midrule
\multirow{4}{*}{1} & Baseline & 0.266 & 5.647 & 6.305 & 0.331 & 0.759 & 0.901 & 0.947 \\
& + NIT                      & 0.190 & 1.824 & 4.848 & 0.257 & 0.763 & 0.919 & 0.965 \\
& + Denoising                & 0.163 & 1.256 & 4.193 & 0.224 & 0.801 & 0.935 & 0.973 \\
& Full Model                  & 0.154 & 1.174 & 4.120 & 0.216 & 0.811 & 0.939 & 0.976 \\
\midrule
\multirow{3}{*}{2} & Baseline & 0.602 & 63.914 & 14.726 & 0.467 & 0.785 & 0.902 & 0.939 \\
& + NIT     & 0.169 & 1.727 & 4.693 & 0.236 & 0.812 & 0.929 & 0.967 \\
& Full Model & 0.131 & 0.926 & 3.731 & 0.188 & 0.852 & 0.949 & 0.980 \\
\bottomrule
\end{tabular}
\vspace{2mm}
\caption{Ablation study showing the importance of different modules in our system. The maximum depth was set to $30$m for this study. `Stride' denotes intra-triplet stride (see supplementary material).}
\label{tab:ablation}
\vspace{-\baselineskip}
\end{table}

\section{Limitations}

Our method has a number of limitations. First, as is typical of monocular methods, it is only able to estimate depth up to scale. Second, like most stereo approaches (whether variable-baseline like ours, or fixed-baseline with a rigid stereo rig), it struggles to preserve the detail of distant parts of the scene because of limited parallax. Third, it also struggles to recover structural detail from very dark image regions (e.g.\ see Figure~\ref{fig:challenges}(b)). And finally, in common with most vision-based depth estimation methods, it does not consistently perform well for transparent surfaces like glass.

\section{Conclusions}
\label{sec:conclusion}

In this paper, we propose a self-supervised method to learn a single model to estimate depth maps from monocular day and nighttime RGB images. By compensating for the illumination changes that can occur from one frame to the next, we enable accurate nighttime depth estimation in non-uniform lighting conditions. Moreover, by predicting per-pixel residual flow and using it to correct the reprojection correspondences induced by the estimated ego-motion and depth, we improve our method's ability to cope with both moving objects in the scene and motion blur. Finally, by denoising the input images prior to calculating the photometric loss, we improve the loss's ability to provide a strong supervision signal, making the entire system more robust and accurate.

\clearpage

\appendix

\part*{Supplementary Material}

\begin{abstract}
In this supplementary material, we provide (i) further details about how we trained and evaluated our networks on the Oxford RobotCar dataset; (ii) a comparison of the requirements of our method vs.\ those of existing state-of-the-art approaches; and (iii) additional experiments and discussion relating to our estimation of ego-motion and lighting changes.
\end{abstract}

\section{Training and Evaluation Details}

\subsection{Dataset Processing}

We evaluated our method's performance on both day and night sequences from the Oxford RobotCar dataset~\cite{RobotCarDatasetIJRR}. This dataset was collected over a one-year period by traversing the same route multiple times so as to include a variety of different weather and lighting conditions. We used the six sequences in the \texttt{2014-12-09-13-21-02} traversal for our daytime experiments, and the six sequences in the \texttt{2014-12-16-18-44-24} traversal for our night-time ones. We cropped the bottom 20\% of each image to remove the car hood, and adjusted the camera intrinsics accordingly. We then filtered each sequence to produce a sub-sequence of keyframes such that each consecutive pair of keyframes was at least $0.5$m apart. We then constructed our training, validation and testing splits from triplets of images in the original sequence, each centred on one keyframe (at this stage with a stride of $1$). This process had the beneficial side-effect of filtering out most of the images that were captured when the ego-vehicle was stationary.

\begin{figure}[!t]
\centering
\subfloat[\centering Splits of RNW~\cite{wang2021regularizing}, ADDS-DepthNet~\cite{liu2021self}]{{\includegraphics[width=7.1cm]{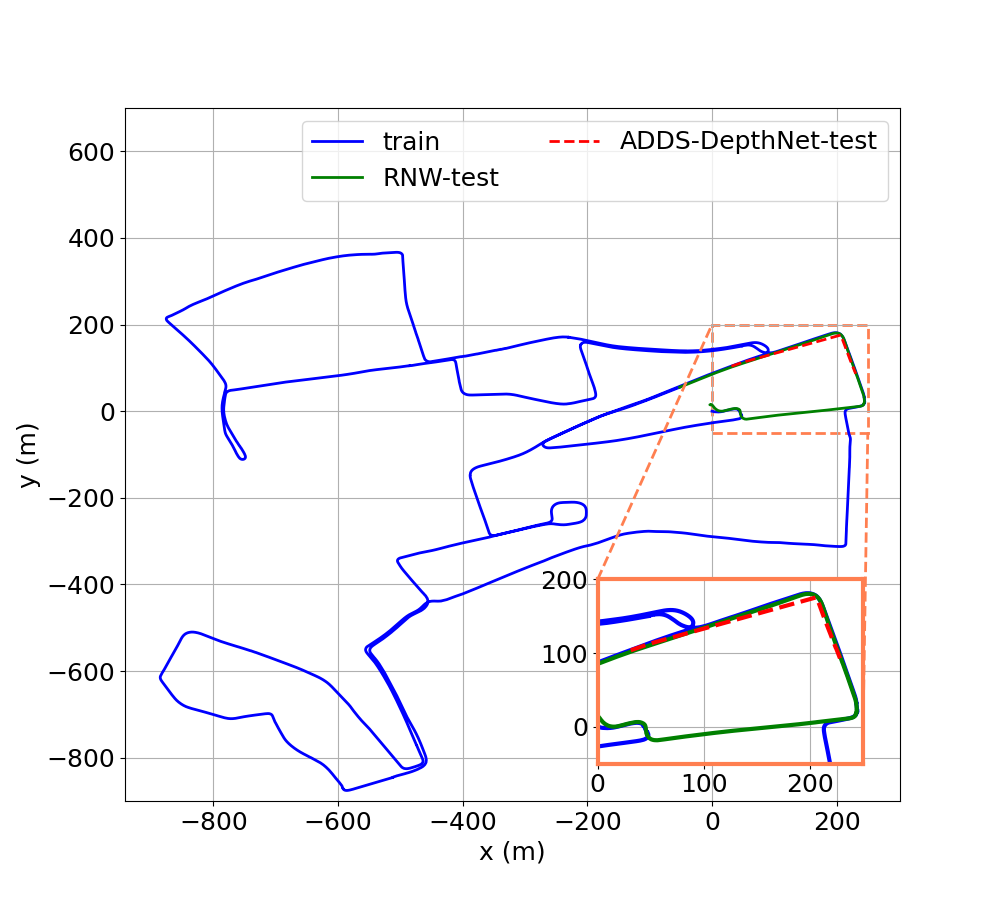} }}%
\subfloat[\centering Our splits]{{\includegraphics[width=7.1cm]{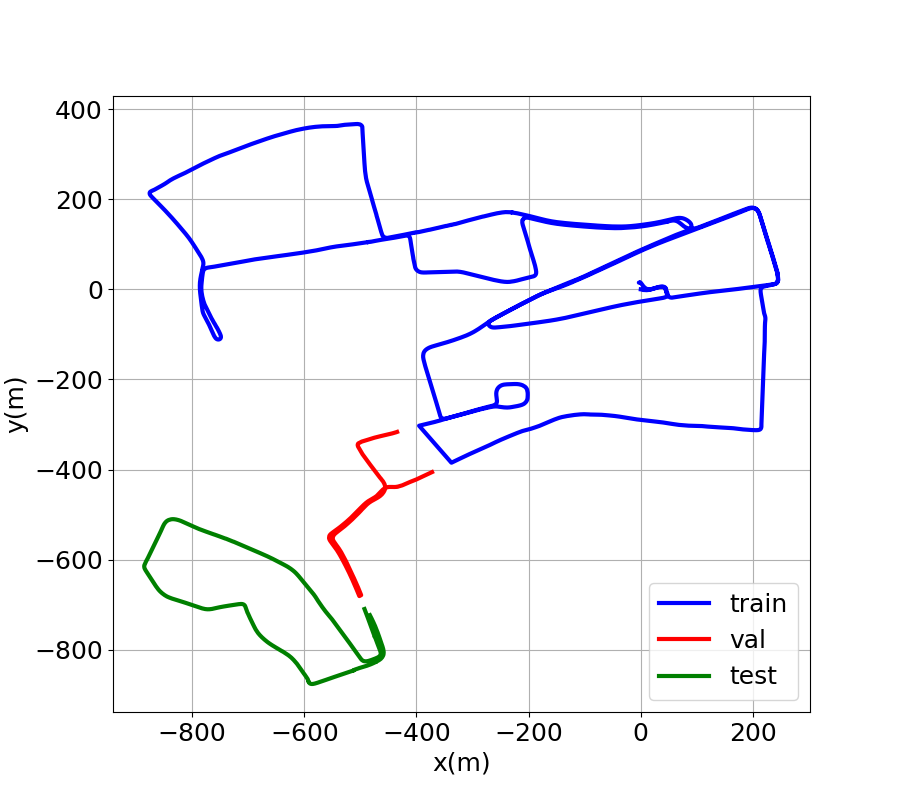} }}%
\caption{A geographic visualisation of the training and testing splits used by (a) RNW~\cite{wang2021regularizing} and ADDS-DepthNet~\cite{liu2021self}, and (b) our method, based on the GPS locations provided with the Oxford RobotCar dataset~\cite{RobotCarDatasetIJRR}. The splits of the methods in (a) clearly suffer from geographical overlap, whereas ours are explicitly constructed to be geographically disjoint, enabling fairer comparisons between methods on this dataset going forwards. (Best viewed in colour.)}
\label{fig:rnw_data_split}
\end{figure}

There are unfortunately no standard splits for the RobotCar dataset, and moreover the splits chosen by other methods~\cite{wang2021regularizing,liu2021self} overlapped geographically (see Figure~\ref{fig:rnw_data_split}(a)). For these reasons, we constructed our own splits by geographically dividing our triplets (constructed as above) into three daytime sets and three nighttime sets (see Figure~\ref{fig:rnw_data_split}(b)). The nighttime splits are comprised of $19,612$ triplets for training,  $2,629$ triplets for validation and $4,559$ triplets for testing. The daytime splits are comprised of $17,790$ triplets for training, $6,693$ triplets for testing and $2,629$ triplets for validation. We used the GPS locations provided with the dataset to ensure that there was no overlap between the training and testing locations in the city. Our hope in constructing these new splits is that they will help enable fairer comparisons between methods on this dataset going forwards.

\subsection{Training}

The networks were trained for $15$ epochs using PyTorch. The learning rate was set to $10^{-4}$. The loss function weights $\lambda_r$ and $\lambda_g$ (see main paper) were both set to $10^{-3}$. We used the popular Adam optimiser with $\beta_1 = 0.9$ and $\beta_2 = 0.99$ to train our network.

\subsection{Evaluation}

For depth evaluation, to avoid redundant computation due to the high camera frame rate, we sub-sampled the test set to have a 2m frame separation, resulting in 709 test images for nighttime and 702 images for daytime. We used the (sparse) LiDAR data provided with the dataset as ground truth for the evaluation, as no official per-pixel ground truth depth is available. As per common practice, we used a script from the Oxford RobotCar SDK to reproject LiDAR points from several nearby frames onto the current test frame to increase the number of points available for evaluation.

\textbf{Scale:} Like other methods trained using only monocular images, our method can only estimate depth maps up to a scale factor. When reporting results, we scale the predictions of all the methods using the median value approach suggested in~\cite{zhou2017unsupervised}.

\textbf{Depth Truncation:} Typically, a depth estimation benchmark will specify one or more maximum distance thresholds, and then, for each such threshold $\tau$, include only pixels whose ground-truth depth is $\le \tau$ in the corresponding depth evaluation. This makes sense, since the intention is to determine the ability of a method to predict depths up to certain maximum distances.

However, for each $\tau$, most (if not all) existing unsupervised depth estimation methods also truncate their \emph{estimated} depth values to $\tau$ (as opposed to e.g.\ counting pixels whose depth is estimated as being $> \tau$ as outliers). Unfortunately, this makes far less sense, and can have the unintended side-effect of biasing the evaluation in their favour. For example, consider a method that estimates a depth $d \gg \tau$ for a pixel $\mathbf{u}$ whose corresponding ground-truth depth is $\tau$. Without truncation, the error for this pixel will be $|d - \tau| \gg 0$, but with truncation, it will be $|\min{}(d,\tau) - \tau| = 0$. In other words, this practice can lead to low errors wrongly being ascribed to mispredicted pixels, which can cause the overall error that is reported to be much lower than it otherwise would have been.

To avoid this problem, and as an alternative to explicitly counting outliers, we truncate all estimated depth values in this paper to a threshold $\tau' \gg \tau$, thus limiting the maximum penalty for each outlier to at most $\tau' - \tau$ (in practice, we set $\tau' = 100$m, which is much larger than $\tau = 50$m, the largest maximum distance for which we report results). The difference this makes to the evaluation can be seen in Table~\ref{tab:max_depth_tab}.

\begin{table}[!t]
\centering
\footnotesize
\begin{tabular}{ccccccccc}
\toprule
$\tau'$ & \textit{Method} & \textit{Abs.\ Rel.} & \textit{Sq.\ Rel.} & \textit{RMSE} & \textit{Log RMSE} & $\delta < 1.25$ & $\delta < 1.25^2$ & $\delta< 1.25^3$ \\
\midrule
\multirow{2}{*}{50m} & RNW~\cite{wang2021regularizing} & 0.180 & 1.525 & 6.183 & 0.259 & 0.740 & 0.913 & 0.962 \\
& Ours & 0.171 & 1.481 & 6.009 & 0.242 & 0.758 & 0.917 & 0.965 \\
\midrule
\multirow{2}{*}{100m} & RNW~\cite{wang2021regularizing} & 0.185 & 1.710 & 6.549 & 0.262 & 0.733 & 0.910 & 0.960 \\
& Ours & 0.174 & 1.637 & 6.302 & 0.245 & 0.754 & 0.915 & 0.964 \\
\bottomrule
\end{tabular}
\vspace{2mm}
\caption{The effects of truncating the depths estimated by different methods to different thresholds, for a maximum evaluation distance of $50$m, evaluated over our test set. The methods' performance appears significantly better when the estimated depths are truncated to $50$m rather than $100$m.}
\label{tab:max_depth_tab}
\vspace{-\baselineskip}
\end{table}

\section{Ease-of-Deployment Comparison with State-of-the-Art Approaches}

To better understand the relative deployability of our method, we compare it to three state-of-the-art approaches, namely DeFeat-Net~\cite{spencer2020defeat}, ADDS-DepthNet~\cite{liu2021self} and RNW~\cite{wang2021regularizing}. Besides outperforming the existing models, our method also has several other advantages over them, including easier data and training requirements, as shown in Table~\ref{tab:requirements}.

\textbf{DeFeat-Net}~\cite{spencer2020defeat} is trained with an additional feature space with pixel-wise contrastive learning, which is assumed to be illumination and weather invariant. This feature network is simultaneously learned with the depth and pose networks. We used their pre-trained model to compare against our method.

\textbf{ADDS-DepthNet}~\cite{liu2021self} is compared against our method using the pre-trained models released alongside their code on GitHub. The orthogonality loss proposed in their method requires corresponding daytime data. It is not possible to train this model with night-only data.

\textbf{RNW}~\cite{wang2021regularizing} did not release their pre-trained models alongside their code. We retrained their model with our training data and reported the results. This method uses a GAN-based regularisation loss, through which they achieve an $\approx 93\%$ improvement over the baseline method. To achieve this, they needed to carefully train a model using daytime images, as the nighttime performance directly depends on how well the daytime model is trained.

The results reported in their paper are trained with a ResNet-50 model, whereas the model we use is ResNet-18. When we tried retraining their approach with ResNet-18, we observed a divergence of training after a few epochs. Training instability is a well-known issue that must be overcome when working with GANs, but this does highlight the relative difficulties of making even simple modifications to a GAN-based system such as RNW in comparison to our own approach. Notably, in spite of such training difficulties, RNW is the only approach that is very close to our method in terms of its error and accuracy metrics. However, unlike our method, their model can only be used for nighttime depth estimation.

\begin{table}[!t]
\centering
\footnotesize
\begin{tabular}{lcccc}
\toprule
\textit{Requirement} & \textit{DeFeat-Net}~\cite{spencer2020defeat} & \textit{ADDS-DepthNet}~\cite{liu2021self} & \textit{RNW}~\cite{wang2021regularizing} & \textit{Ours} \\
\midrule
No additional features     & \xmark & \cmark & \cmark & \cmark \\
No daytime images          & \cmark & \xmark & \xmark & \cmark \\
No paired day-night images & \cmark & \xmark & \cmark & \cmark \\
No GAN losses              & \cmark & \cmark & \xmark & \cmark \\
Night-only training        & \cmark & \xmark & \xmark & \cmark \\
Works for day and night    & \cmark & \cmark & \xmark & \cmark \\
\bottomrule
\end{tabular}
\vspace{2mm}
\caption{Ease-of-deployment comparison of our method with three state-of-the-art approaches}
\label{tab:requirements}
\vspace{-\baselineskip}
\end{table}

\section{Ego-Motion Estimation}

\begin{figure}[!t]
\centering
\includegraphics[scale=0.4]{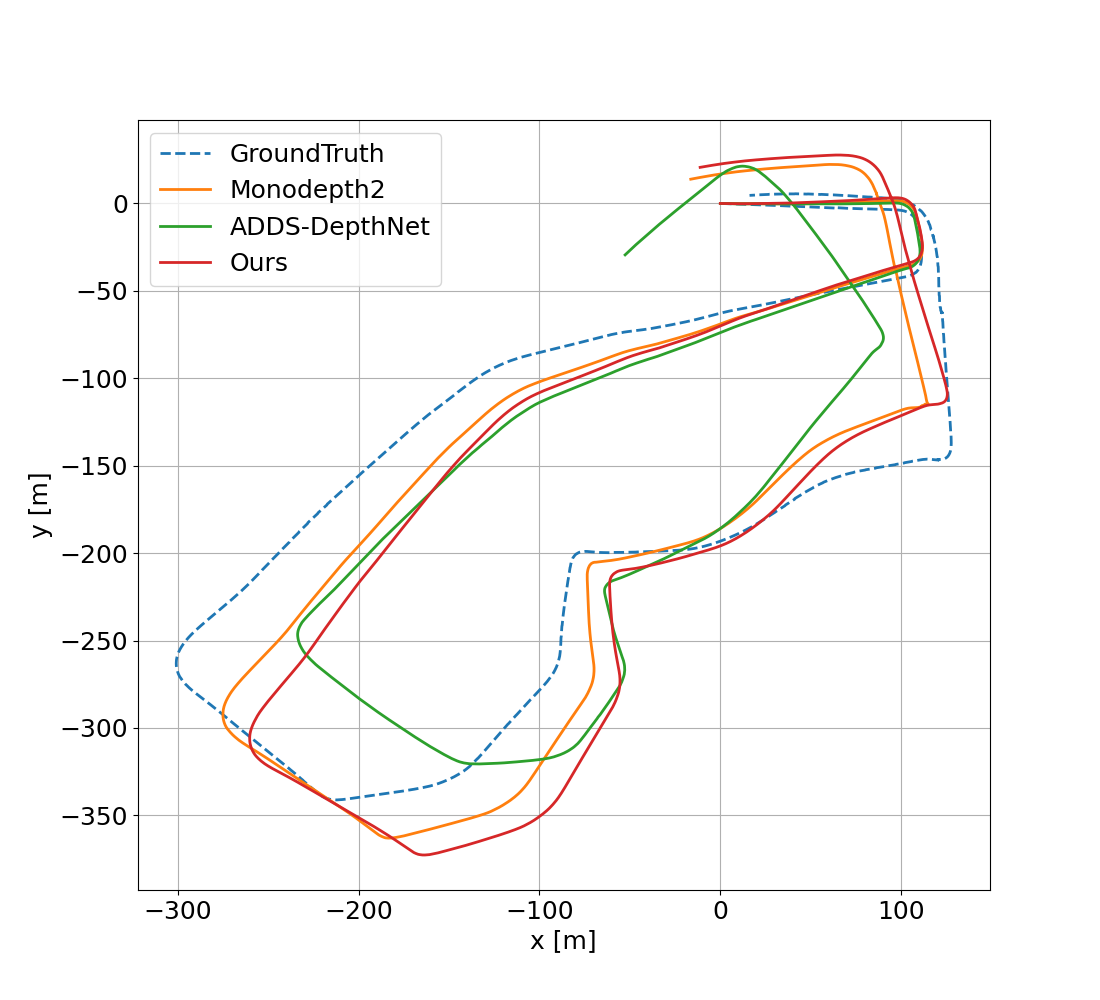}
\caption{Comparing the trajectory estimated by our method for our nighttime test sequence to the trajectories estimated by Monodepth2~\cite{godard2018digging} and ADDS-DepthNet~\cite{liu2021self}, and to the ground truth.}
\label{fig:pose_results}
\end{figure}

To evaluate the ability of our method to estimate the ego-motion of the camera, we compare the absolute trajectory error (ATE)~\cite{zhou2017unsupervised,vankadari2019unsupervised} it can achieve on our nighttime test sequence to the ATEs of three state-of-the-art methods, namely Monodepth2~\cite{godard2018digging}, DeFeat-Net~\cite{spencer2020defeat} and ADDS-DepthNet~\cite{liu2021self} (see Table~\ref{tab:pose_results}). We also visualise the overall trajectory we estimate and compare it to those of other methods, as well as the ground truth (see Figure~\ref{fig:pose_results}). Note that we estimate the ego-motion only up to scale, as our method uses monocular images during training. Moreover, the scale can drift over the course of the sequence, as there is no external constraint enforced to keep it constant.

We used the rescaling approach described in~\cite{zhou2017unsupervised} for both our quantitative and qualitative pose estimation results. In both cases, our proposed method achieves results that are competitive with the state-of-the-art.

\begin{table}[!t]
\centering
\footnotesize
\begin{tabular}{lcc}
\toprule
\textit{Method} & $t_{ate}$ $(m) \downarrow$ & $r_{ate}$ $(rad) \downarrow$   \\
\midrule
Monodepth2 \cite{godard2018digging} & 0.01317 $\pm$ 0.01517 & \textbf{0.00036} $\pm$ 0.00040 \\
DeFeat-Net~\cite{spencer2020defeat} & 0.03760 $\pm$ 0.01990 & 0.00115 $\pm$ 0.00129 \\
ADDS-DepthNet~\cite{liu2021self} & 0.01312 $\pm$ 0.01386 & 0.00083 $\pm$ 0.00061 \\
Ours & \textbf{0.01310} $\pm$ 0.01348 & 0.00041 $\pm$ 0.00039 \\
\bottomrule
\end{tabular}
\vspace{2mm}
\caption{A quantitative comparison of our ATE errors to those of Monodepth2~\cite{godard2018digging}, DeFeat-Net~\cite{spencer2020defeat} and ADDS-DepthNet~\cite{liu2021self} on our nighttime test sequence. (The results are in the form mean $\pm$ std.)}
\label{tab:pose_results}
\vspace{-2\baselineskip}
\end{table}

\section{Lighting Change Compensation}

In this section, we aim to provide a greater understanding of the lighting change compensation technique we propose in the main paper by motivating it from first principles.

Recall that at frame $t$, our approach first predicts two per-pixel change images, $C_t$ and $B_t$, that aim to capture the per-pixel changes in contrast (scale) and brightness (shift) that occur between frame $t$ and frame $t+1$. It then uses these change images to transform the reconstructed image $I'_t$ via $\tilde{I}_t = C_t \odot I'_t + B_t$ before it is compared with image $I_t$.

A natural question to ask about this is: `why is this linear transformation an appropriate model of the lighting changes between frames $t$ and $t+1$?' In this section, we will attempt to provide an answer.

\subsection{Theoretical Background}

\textbf{Image Formation:} Recall the basic Phong illumination model from Computer Graphics, which calculates the brightness at a 3D point $\mathbf{p}$ in world space when viewed from a direction $-\mathbf{\hat{V}}(\mathbf{p})$:
\begin{equation}
B(\mathbf{p}) = B_a k_a + k_d \sum_{l \in \mathit{lights}} B_l \left( (\mathbf{\hat{N}}(\mathbf{p}) \cdot \mathbf{\hat{L}}_l(\mathbf{p})) + k_s(\mathbf{\hat{R}}_l(\mathbf{p}) \cdot \mathbf{\hat{V}}(\mathbf{p}))^{k_n} \right).
\end{equation}
In this, as usual, $B_a$ denotes the ambient illumination, $B_l$ denotes the brightness of light $l$, $\{k_a,k_d,k_s\}$ denote the \{ambient,diffuse,specular\} reflection coefficients and $k_n$ the specular exponent associated with the surface material, $\mathbf{\hat{N}}(\mathbf{p})$ denotes the unit normal to the surface at $\mathbf{p}$, $\mathbf{\hat{L}}_l(\mathbf{p})$ denotes a unit vector from $\mathbf{p}$ towards $\mathbf{p}_l$, the position of light $l$, $\mathbf{\hat{R}}_l(\mathbf{p})$ denotes the result of reflecting $\mathbf{\hat{L}}_l(\mathbf{p})$ across $\mathbf{N}(\mathbf{p})$, and $\mathbf{\hat{V}}(\mathbf{p})$ denotes a unit vector from $\mathbf{p}$ towards the viewer.

The image formation process can then be modelled as
\begin{equation}
I(\mathbf{u}) = f(e\mathcal{V}(\mathbf{u})B(\mathbf{p})),
\end{equation}
in which $\mathbf{u}$ is the pixel corresponding to $\mathbf{p}$ on the image plane, $f$ denotes the camera response function, $e$ denotes the exposure time, and $\mathcal{V}$ denotes the Vignetting function. This is commonly simplified (e.g.~\cite{jin2001real, yang2020d3vo}) by assuming that $f$ is the identity function and that there is no Vignetting effect, in which case the model simplifies to
\begin{equation}
I(\mathbf{u}) = eB(\mathbf{p}).
\end{equation}
\textbf{Further Simplification:} For our purposes here, we will simplify this model still further by assuming that the only illumination in the scene is diffuse, which allows us to approximate $B(\mathbf{p})$ as
\begin{equation}
B(\mathbf{p}) \approx k_d \sum_{l \in \mathit{lights}} B_l (\mathbf{\hat{N}}(\mathbf{p}) \cdot \mathbf{\hat{L}}_l(\mathbf{p})).
\end{equation}
Whilst this assumption is not really true in practice (though the \emph{ambient} illumination at night is likely to be negligible), it will significantly simplify our derivation in what follows.

\textbf{Lighting Change Model:} To model the change in lighting between frames $t$ and $t+1$ for a 3D world-space point $\mathbf{p}$ that is imaged at pixels $\mathbf{u}$ in $I_t$ and $\mathbf{u}'$ in $I_{t+1}$, we can first approximate $I_t(\mathbf{u})$ and $I_{t+1}(\mathbf{u'})$ using our image formation model:
\begin{equation}
\begin{aligned}
I_t(\mathbf{u}) &\approx e^{(t)} \; k_d \sum_{l \in \mathit{lights}} B_l^{(t)}(\mathbf{\hat{N}}(\mathbf{p}) \cdot \mathbf{\hat{L}}_l^{(t)}(\mathbf{p})) \\
I_{t+1}(\mathbf{u'}) &\approx e^{(t+1)} \; k_d \sum_{l \in \mathit{lights}} B_l^{(t+1)}(\mathbf{\hat{N}}(\mathbf{p}) \cdot \mathbf{\hat{L}}_l^{(t+1)}(\mathbf{p})).
\end{aligned}
\end{equation}
In this, $e^{(t)}$ denotes the exposure time at frame $t$, $B_l^{(t)}$ denotes the brightness of light $l$ at frame $t$ and $\mathbf{\hat{L}}_l^{(t)}(\mathbf{p})$ denotes a unit vector from $\mathbf{p}$ towards $\mathbf{p}_l^{(t)}$, the position of light $l$ at frame $t$. Note that both the brightness and the position of each light in the scene are now assumed to potentially vary with time (e.g.\ a car headlight will move with the car).

We next observe that both $I_t$ and $I_{t+1}$ are actually 3-channel RGB images, not single-channel intensity images as in the simplest version of the Phong illumination model. In Computer Graphics, this is commonly handled by assigning a 3-channel RGB colour to each light, thus making each $B_l^{(t)}$ a 3-channel vector. However, it is more physically correct to think of the diffuse reflection coefficient $k_d$ as a 3-channel vector that captures the extent to which different components of white light get reflected/absorbed by the surface material. Moreover, it will turn out to be possible to eliminate $k_d$ in this case, and so for our purposes here, we will take this latter view.

To eliminate $k_d$, it suffices to divide $I_t(\mathbf{u})$ by $I_{t+1}(\mathbf{u'})$ via
\begin{equation}
I_t(\mathbf{u}) \oslash I_{t+1}(\mathbf{u}) \approx \frac{e^{(t)} \sum_{l \in \mathit{lights}} B_l^{(t)}(\mathbf{\hat{N}}(\mathbf{p}) \cdot \mathbf{\hat{L}}_l^{(t)}(\mathbf{p}))}{e^{(t+1)} \sum_{l \in \mathit{lights}} B_l^{(t+1)}(\mathbf{\hat{N}}(\mathbf{p}) \cdot \mathbf{\hat{L}}_l^{(t+1)}(\mathbf{p}))} \mathbbm{1} = C_t^\star(\mathbf{u}) \mathbbm{1},
\end{equation}
in which $\oslash$ denotes Hadamard division and $\mathbbm{1} = (1,1,1)^\top \in \mathbb{R}^3$. This yields a scalar, $C_t^\star(\mathbf{u})$, that can clearly be multiplied by $I_{t+1}(\mathbf{u'})$ to approximate $I_t(\mathbf{u})$ via $C_t^\star(\mathbf{u}) I_{t+1}(\mathbf{u'}) \approx I_t(\mathbf{u})$. Interestingly, since $I'_t(\mathbf{u}) \approx I_{t+1}(\mathbf{u'})$ as per the definition in the main paper, this means that $C_t^\star(\mathbf{u}) I'_t(\mathbf{u}) \approx I_t(\mathbf{u})$, which we can also write as
\begin{equation}
I_t \approx C_t^\star \odot I'_t.
\label{eqn:theoreticalmodel}
\end{equation}
This corresponds to a single-channel lighting change model in which we will aim to compensate for the lighting changes in $I'_t$ by simply multiplying the value of each pixel in $I'_t$ by its corresponding scaling factor.

\subsection{Scale-only Model}
\label{subsec:scaleonlymodel}

\begin{figure}[!t]
\centering
\includegraphics[scale=0.3]{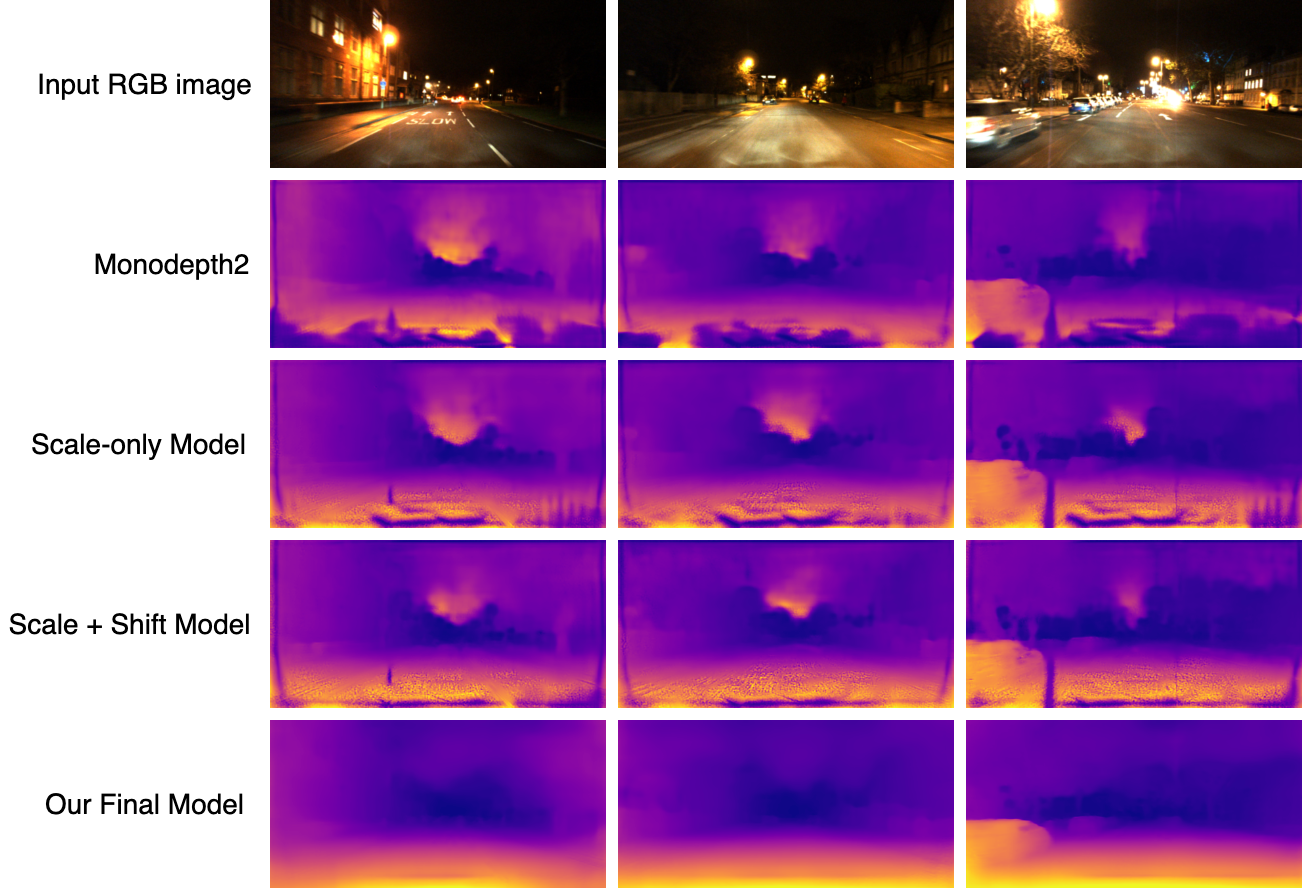}
\caption{A qualitative comparison showing the differences in depth estimation performance between Monodepth2~\cite{godard2018digging}, our scale-only lighting change model (see \S\ref{subsec:scaleonlymodel}) and our scale + shift lighting change model (see \S\ref{subsec:scaleshiftmodel}). Our final model (which additionally includes the motion compensation and denoising components) achieves the best overall performance.}
\label{fig:my_label}
\end{figure}

Based on the theoretical formulation in the previous sub-section, it makes sense to try to predict a single per-pixel change image $C_t$ for each frame $t$ (using a single-channel variant of the lighting change decoder described in the main paper) that can take the place of $C_t^\star$ in Equation~\ref{eqn:theoreticalmodel} and be used to multiply $I'_t$ to compensate for lighting changes. The effects of applying such a scale-only lighting change compensation technique to Monodepth2~\cite{godard2018digging} can be seen in Figure~\ref{fig:my_label}, in which it can be clearly seen that e.g.\ the sizes of the holes directly in front of the car have been significantly reduced. Noticeably, however, our scale-only model still struggles to cope with the strong intensity discontinuities caused by the car headlights in the close-range image region.

\subsection{Scale + Shift Model}
\label{subsec:scaleshiftmodel}

Since the scale-only model is a purely linear intensity transform, an additive term in the form of a shift channel $B_t$ enables an affine transformation, which provides an additional degree of freedom to the network for learning a suitable intensity transformation per pixel. As shown in Figure~\ref{fig:my_label}, this results in significant improvements in the estimated depth for the affected areas. We observed that our affine (scale+shift) model specifically improved depth estimation for pixels close to the camera that represent textureless road patches illuminated non-uniformly by the ego-vehicle's headlights. We hypothesise that given the specific nature of this image region, the affine model benefits from additional learnable parameters.

\acknowledgments{This work was supported by Amazon Web Services via the Oxford-Singapore Human-Machine Collaboration Programme.}

\bibliography{arxiv}

\begin{thebibliography}{52}
\providecommand{\natexlab}[1]{#1}
\providecommand{\url}[1]{\texttt{#1}}
\expandafter\ifx\csname urlstyle\endcsname\relax
  \providecommand{\doi}[1]{doi: #1}\else
  \providecommand{\doi}{doi: \begingroup \urlstyle{rm}\Url}\fi

\bibitem[Yurtsever et~al.(2020)Yurtsever, Lambert, Carballo, and
  Takeda]{yurtsever2020survey}
E.~Yurtsever, J.~Lambert, A.~Carballo, and K.~Takeda.
\newblock {A Survey of Autonomous Driving: Common Practices and Emerging
  Technologies}.
\newblock \emph{IEEE Access}, 8:\penalty0 58443--58469, 2020.

\bibitem[Sanchez et~al.(2018)Sanchez, Corrales, Bouzgarrou, and
  Mezouar]{sanchez2018robotic}
J.~Sanchez, J.-A. Corrales, B.-C. Bouzgarrou, and Y.~Mezouar.
\newblock Robotic manipulation and sensing of deformable objects in domestic
  and industrial applications: a survey.
\newblock \emph{{IJRR}}, 37\penalty0 (7):\penalty0 688--716, 2018.

\bibitem[Livingston et~al.(2009)Livingston, Ai, Swan, and
  Smallman]{livingston2009indoor}
M.~A. Livingston, Z.~Ai, J.~E. Swan, and H.~S. Smallman.
\newblock {Indoor vs.\ Outdoor Depth Perception for Mobile Augmented Reality}.
\newblock In \emph{IEEE Virtual Reality Conference}, pages 55--62, 2009.

\bibitem[Newcombe et~al.(2011)Newcombe, Lovegrove, and
  Davison]{newcombe2011dtam}
R.~A. Newcombe, S.~J. Lovegrove, and A.~J. Davison.
\newblock {DTAM: Dense Tracking and Mapping in Real-Time}.
\newblock In \emph{{ICCV}}, pages 2320--2327, 2011.

\bibitem[Wang and Shen(2018)]{wang2018mvdepthnet}
K.~Wang and S.~Shen.
\newblock {MVDepthNet: Real-time Multiview Depth Estimation Neural Network}.
\newblock In \emph{{3DV}}, pages 248--257, 2018.

\bibitem[Eigen et~al.(2014)Eigen, Puhrsch, and Fergus]{eigen2014depth}
D.~Eigen, C.~Puhrsch, and R.~Fergus.
\newblock {Depth Map Prediction from a Single Image using a Multi-Scale Deep
  Network}.
\newblock In \emph{{NeurIPS}}, pages 2366--2374, 2014.

\bibitem[Liu et~al.(2015)Liu, Shen, and Lin]{liu2015deep}
F.~Liu, C.~Shen, and G.~Lin.
\newblock {Deep Convolutional Neural Fields for Depth Estimation from a Single
  Image}.
\newblock In \emph{{CVPR}}, pages 5162--5170, 2015.

\bibitem[Garg et~al.(2016)Garg, BG, Carneiro, and Reid]{garg2016unsupervised}
R.~Garg, V.~K. BG, G.~Carneiro, and I.~Reid.
\newblock {Unsupervised CNN for Single View Depth Estimation: Geometry to the
  Rescue}.
\newblock In \emph{{ECCV}}, pages 740--756, 2016.

\bibitem[Zhan et~al.(2018)Zhan, Garg, Weerasekera, Li, Agarwal, and
  Reid]{zhan2018unsupervised}
H.~Zhan, R.~Garg, C.~S. Weerasekera, K.~Li, H.~Agarwal, and I.~Reid.
\newblock {Unsupervised Learning of Monocular Depth Estimation and Visual
  Odometry with Deep Feature Reconstruction}.
\newblock In \emph{{CVPR}}, pages 340--349, 2018.

\bibitem[Almalioglu et~al.(2019)Almalioglu, Saputra, de~Gusm{\~a}o, Markham,
  and Trigoni]{almalioglu2018ganvo}
Y.~Almalioglu, M.~R.~U. Saputra, P.~P.~B. de~Gusm{\~a}o, A.~Markham, and
  N.~Trigoni.
\newblock {GANVO: Unsupervised Deep Monocular Visual Odometry and Depth
  Estimation with Generative Adversarial Networks}.
\newblock In \emph{{ICRA}}, pages 5474--5480, 2019.

\bibitem[Kerl et~al.(2013{\natexlab{a}})Kerl, Sturm, and
  Cremers]{kerl2013dense}
C.~Kerl, J.~Sturm, and D.~Cremers.
\newblock {Dense Visual SLAM for RGB-D Cameras}.
\newblock In \emph{{IROS}}, pages 2100--2106, 2013{\natexlab{a}}.

\bibitem[Kerl et~al.(2013{\natexlab{b}})Kerl, Sturm, and
  Cremers]{kerl2013robust}
C.~Kerl, J.~Sturm, and D.~Cremers.
\newblock {Robust Odometry Estimation for RGB-D Cameras}.
\newblock In \emph{{ICRA}}, pages 3748--3754, 2013{\natexlab{b}}.

\bibitem[Comport et~al.(2007)Comport, Malis, and Rives]{comport2007accurate}
A.~I. Comport, E.~Malis, and P.~Rives.
\newblock {Accurate Quadrifocal Tracking for Robust 3D Visual Odometry}.
\newblock In \emph{{ICRA}}, pages 40--45, 2007.

\bibitem[Sharma et~al.(2020)Sharma, Cheong, Heng, and Tan]{sharma2020nighttime}
A.~Sharma, L.-F. Cheong, L.~Heng, and R.~T. Tan.
\newblock {Nighttime Stereo Depth Estimation using Joint Translation-Stereo
  Learning: Light Effects and Uninformative Regions}.
\newblock In \emph{{3DV}}, pages 23--31, 2020.

\bibitem[Portz et~al.(2012)Portz, Zhang, and Jiang]{portz2012optical}
T.~Portz, L.~Zhang, and H.~Jiang.
\newblock {Optical Flow in the Presence of Spatially-Varying Motion Blur}.
\newblock In \emph{{CVPR}}, pages 1752--1759, 2012.

\bibitem[Rav-Acha and Peleg(2000)]{rav2000restoration}
A.~Rav-Acha and S.~Peleg.
\newblock {Restoration of Multiple Images with Motion Blur in Different
  Directions}.
\newblock In \emph{Proceedings of the Fifth IEEE Workshop on Applications of
  Computer Vision}, pages 22--28, 2000.

\bibitem[Huang et~al.(2021)Huang, Li, Jia, Lu, and
  Liu]{huang2021neighbor2neighbor}
T.~Huang, S.~Li, X.~Jia, H.~Lu, and J.~Liu.
\newblock {Neighbor2Neighbor: Self-Supervised Denoising from Single Noisy
  Images}.
\newblock In \emph{{CVPR}}, pages 14781--14790, 2021.

\bibitem[Scharstein and Szeliski(2002)]{scharstein2002taxonomy}
D.~Scharstein and R.~Szeliski.
\newblock {A Taxonomy and Evaluation of Dense Two-Frame Stereo Correspondence
  Algorithms}.
\newblock \emph{{IJCV}}, 47\penalty0 (1-3):\penalty0 7--42, 2002.

\bibitem[Scharstein and Pal(2007)]{scharstein2007learning}
D.~Scharstein and C.~Pal.
\newblock {Learning Conditional Random Fields for Stereo}.
\newblock In \emph{{CVPR}}, pages 1--8, 2007.

\bibitem[Zou and Li(2010)]{zou2010method}
L.~Zou and Y.~Li.
\newblock {A Method of Stereo Vision Matching Based on OpenCV}.
\newblock In \emph{International Conference on Audio, Language and Image
  Processing}, pages 185--190, 2010.

\bibitem[Sch{\"o}nberger and Frahm(2016)]{schonberger2016structure}
J.~L. Sch{\"o}nberger and J.-M. Frahm.
\newblock {Structure-from-Motion Revisited}.
\newblock In \emph{{CVPR}}, pages 4104--4113, 2016.

\bibitem[Dai et~al.(2013)Dai, Li, and He]{dai2013projective}
Y.~Dai, H.~Li, and M.~He.
\newblock {Projective Multiview Structure and Motion from Element-Wise
  Factorization}.
\newblock \emph{{TPAMI}}, 35\penalty0 (9):\penalty0 2238--2251, 2013.

\bibitem[Yu and Gallup(2014)]{yu20143d}
F.~Yu and D.~Gallup.
\newblock {3D Reconstruction from Accidental Motion}.
\newblock In \emph{{CVPR}}, pages 3986--3993, 2014.

\bibitem[Ladick{\'y} et~al.(2014)Ladick{\'y}, Shi, and
  Pollefeys]{ladicky2014pulling}
L.~Ladick{\'y}, J.~Shi, and M.~Pollefeys.
\newblock {Pulling Things out of Perspective}.
\newblock In \emph{{CVPR}}, pages 89--96, 2014.

\bibitem[dos Santos~Rosa et~al.(2019)dos Santos~Rosa, Guizilini, and
  Grassi~Jr]{dos2019sparse}
N.~dos Santos~Rosa, V.~Guizilini, and V.~Grassi~Jr.
\newblock {Sparse-to-Continuous: Enhancing Monocular Depth Estimation using
  Occupancy Maps}.
\newblock In \emph{International Conference on Advanced Robotics}, pages
  793--800, 2019.

\bibitem[Fu et~al.(2018)Fu, Gong, Wang, Batmanghelich, and Tao]{fu2018deep}
H.~Fu, M.~Gong, C.~Wang, K.~Batmanghelich, and D.~Tao.
\newblock {Deep Ordinal Regression Network for Monocular Depth Estimation}.
\newblock In \emph{{CVPR}}, pages 2002--2011, 2018.

\bibitem[Godard et~al.(2017)Godard, Mac~Aodha, and
  Brostow]{godard2017unsupervised}
C.~Godard, O.~Mac~Aodha, and G.~J. Brostow.
\newblock {Unsupervised Monocular Depth Estimation with Left-Right
  Consistency}.
\newblock In \emph{{CVPR}}, pages 6602--6611, 2017.

\bibitem[Jaderberg et~al.(2015)Jaderberg, Simonyan, Zisserman, and
  Kavukcuoglu]{jaderberg2015spatial}
M.~Jaderberg, K.~Simonyan, A.~Zisserman, and K.~Kavukcuoglu.
\newblock {Spatial Transformer Networks}.
\newblock In \emph{{NeurIPS}}, pages 2017--2025, 2015.

\bibitem[Wang et~al.(2004)Wang, Bovik, Sheikh, and Simoncelli]{wang2004image}
Z.~Wang, A.~C. Bovik, H.~R. Sheikh, and E.~P. Simoncelli.
\newblock {Image Quality Assessment: From Error Visibility to Structural
  Similarity}.
\newblock \emph{{TIP}}, 13\penalty0 (4):\penalty0 600--612, 2004.

\bibitem[Zhou et~al.(2017)Zhou, Brown, Snavely, and Lowe]{zhou2017unsupervised}
T.~Zhou, M.~Brown, N.~Snavely, and D.~G. Lowe.
\newblock {Unsupervised Learning of Depth and Ego-Motion from Video}.
\newblock In \emph{{CVPR}}, pages 1851--1860, 2017.

\bibitem[Vankadari et~al.(2018)Vankadari, Das, Majumdar, and
  Kumar]{babu2018undemon}
M.~B. Vankadari, K.~Das, A.~Majumdar, and S.~Kumar.
\newblock {UnDEMoN: Unsupervised Deep Network for Depth and Ego-Motion
  Estimation}.
\newblock In \emph{{IROS}}, pages 1082--1088, 2018.

\bibitem[Li et~al.(2018)Li, Wang, Long, and Gu]{li2018undeepvo}
R.~Li, S.~Wang, Z.~Long, and D.~Gu.
\newblock {UnDeepVO: Monocular Visual Odometry through Unsupervised Deep
  Learning}.
\newblock In \emph{{ICRA}}, pages 7286--7291, 2018.

\bibitem[Yin and Shi(2018)]{yin2018geonet}
Z.~Yin and J.~Shi.
\newblock {GeoNet: Unsupervised Learning of Dense Depth, Optical Flow and
  Camera Pose}.
\newblock In \emph{{CVPR}}, pages 1983--1992, 2018.

\bibitem[Luo et~al.(2020)Luo, Yang, Wang, Wang, Xu, Nevatia, and
  Yuille]{luo2018every}
C.~Luo, Z.~Yang, P.~Wang, Y.~Wang, W.~Xu, R.~Nevatia, and A.~Yuille.
\newblock {Every Pixel Counts++: Joint Learning of Geometry and Motion with 3D
  Holistic Understanding}.
\newblock \emph{{TPAMI}}, 42\penalty0 (10):\penalty0 2624--2641, 2020.

\bibitem[Aleotti et~al.(2018)Aleotti, Tosi, Poggi, and
  Mattoccia]{aleotti2018generative}
F.~Aleotti, F.~Tosi, M.~Poggi, and S.~Mattoccia.
\newblock {Generative Adversarial Networks for unsupervised monocular depth
  prediction}.
\newblock In \emph{{ECCV-W}}, 2018.

\bibitem[Vankadari et~al.(2019)Vankadari, Kumar, Majumder, and
  Das]{vankadari2019unsupervised}
M.~Vankadari, S.~Kumar, A.~Majumder, and K.~Das.
\newblock {Unsupervised Learning of Monocular Depth and Ego-Motion using
  Conditional PatchGANs}.
\newblock In \emph{{IJCAI}}, pages 5677--5684, 2019.

\bibitem[Godard et~al.(2019)Godard, Mac~Aodha, Firman, and
  Brostow]{godard2018digging}
C.~Godard, O.~Mac~Aodha, M.~Firman, and G.~Brostow.
\newblock {Digging Into Self-Supervised Monocular Depth Estimation}.
\newblock In \emph{{ICCV}}, pages 3828--3838, 2019.

\bibitem[Lyu et~al.(2021)Lyu, Liu, Wang, Kong, Liu, Liu, Chen, and
  Yuan]{lyu2021hr}
X.~Lyu, L.~Liu, M.~Wang, X.~Kong, L.~Liu, Y.~Liu, X.~Chen, and Y.~Yuan.
\newblock {HR-Depth: High Resolution Self-Supervised Monocular Depth
  Estimation}.
\newblock In \emph{{AAAI}}, pages 2294--2301, 2021.

\bibitem[Petrovai and Nedevschi(2022)]{petrovai2022exploiting}
A.~Petrovai and S.~Nedevschi.
\newblock {Exploiting Pseudo Labels in a Self-Supervised Learning Framework for
  Improved Monocular Depth Estimation}.
\newblock In \emph{{CVPR}}, pages 1578--1588, 2022.

\bibitem[Yan et~al.(2021)Yan, Zhao, Bu, and Jin]{yan2021channel}
J.~Yan, H.~Zhao, P.~Bu, and Y.~Jin.
\newblock {Channel-Wise Attention-Based Network for Self-Supervised Monocular
  Depth Estimation}.
\newblock In \emph{{3DV}}, pages 464--473, 2021.

\bibitem[Hui(2022)]{hui2022rm}
T.-W. Hui.
\newblock {RM-Depth: Unsupervised Learning of Recurrent Monocular Depth in
  Dynamic Scenes}.
\newblock In \emph{{CVPR}}, pages 1675--1684, 2022.

\bibitem[Guizilini et~al.(2020)Guizilini, Ambru\cb{s}, Pillai, Raventos, and
  Gaidon]{guizilini20203d}
V.~Guizilini, R.~Ambru\cb{s}, S.~Pillai, A.~Raventos, and A.~Gaidon.
\newblock {3D Packing for Self-Supervised Monocular Depth Estimation}.
\newblock In \emph{{CVPR}}, pages 2485--2494, 2020.

\bibitem[Shu et~al.(2020)Shu, Yu, Duan, and Yang]{shu2020feature}
C.~Shu, K.~Yu, Z.~Duan, and K.~Yang.
\newblock {Feature-metric Loss for Self-supervised Learning of Depth and
  Egomotion}.
\newblock In \emph{{ECCV}}, pages 572--588, 2020.

\bibitem[Li et~al.(2021)Li, Gordon, Zhao, Casser, and
  Angelova]{li2020unsupervised}
H.~Li, A.~Gordon, H.~Zhao, V.~Casser, and A.~Angelova.
\newblock {Unsupervised Monocular Depth Learning in Dynamic Scenes}.
\newblock \emph{{PMLR}}, 155:\penalty0 1908--1917, 2021.

\bibitem[Spencer et~al.(2020)Spencer, Bowden, and Hadfield]{spencer2020defeat}
J.~Spencer, R.~Bowden, and S.~Hadfield.
\newblock {DeFeat-Net: General Monocular Depth via Simultaneous Unsupervised
  Representation Learning}.
\newblock In \emph{{CVPR}}, pages 14402--14413, 2020.

\bibitem[Vankadari et~al.(2020)Vankadari, Garg, Majumder, Kumar, and
  Behera]{vankadari2020unsupervised}
M.~Vankadari, S.~Garg, A.~Majumder, S.~Kumar, and A.~Behera.
\newblock {Unsupervised Monocular Depth Estimation for Night-time Images using
  Adversarial Domain Feature Adaptation}.
\newblock In \emph{{ECCV}}, pages 443--459, 2020.

\bibitem[Wang* et~al.(2021)Wang*, Zhang*, Yan, Li, Xu, Li, and
  Yang]{wang2021regularizing}
K.~Wang*, Z.~Zhang*, Z.~Yan, X.~Li, B.~Xu, J.~Li, and J.~Yang.
\newblock {Regularizing Nighttime Weirdness: Efficient Self-supervised
  Monocular Depth Estimation in the Dark}.
\newblock In \emph{{ICCV}}, pages 16055--16064, 2021.

\bibitem[Liu et~al.(2021)Liu, Song, Wang, Liu, and Zhang]{liu2021self}
L.~Liu, X.~Song, M.~Wang, Y.~Liu, and L.~Zhang.
\newblock {Self-supervised Monocular Depth Estimation for All Day Images using
  Domain Separation}.
\newblock In \emph{{ICCV}}, pages 12737--12746, 2021.

\bibitem[Jin et~al.(2001)Jin, Favaro, and Soatto]{jin2001real}
H.~Jin, P.~Favaro, and S.~Soatto.
\newblock {Real-Time Feature Tracking and Outlier Rejection with Changes in
  Illumination}.
\newblock In \emph{{ICCV}}, volume~1, pages 684--689, 2001.

\bibitem[Baker and Matthews(2004)]{baker2004lucas}
S.~Baker and I.~Matthews.
\newblock {Lucas-Kanade 20 Years On: A Unifying Framework}.
\newblock \emph{{IJCV}}, 56\penalty0 (3):\penalty0 221--255, 2004.

\bibitem[Yang et~al.(2020)Yang, Stumberg, Wang, and Cremers]{yang2020d3vo}
N.~Yang, L.~v. Stumberg, R.~Wang, and D.~Cremers.
\newblock {D3VO: Deep Depth, Deep Pose and Deep Uncertainty for Monocular
  Visual Odometry}.
\newblock In \emph{{CVPR}}, pages 1281--1292, 2020.

\bibitem[Maddern et~al.(2017)Maddern, Pascoe, Linegar, and
  Newman]{RobotCarDatasetIJRR}
W.~Maddern, G.~Pascoe, C.~Linegar, and P.~Newman.
\newblock {1 year, 1000 km: The Oxford RobotCar Dataset}.
\newblock \emph{{IJRR}}, 36\penalty0 (1):\penalty0 3--15, 2017.

\end{thebibliography}

\end{document}